\documentclass[conference]{IEEEtran}
\IEEEoverridecommandlockouts

\usepackage{cite}
\usepackage{amsmath,amssymb,amsfonts}
\usepackage{graphicx}
\usepackage{textcomp}
\usepackage{xcolor}

\usepackage{colortbl}
\usepackage{algorithm,algorithmic}
\usepackage{xurl}
\usepackage{makecell}
\usepackage{multirow}  
\usepackage{booktabs}  
\usepackage{cancel}

\def\BibTeX{{\rm B\kern-.05em{\sc i\kern-.025em b}\kern-.08em
    T\kern-.1667em\lower.7ex\hbox{E}\kern-.125emX}}
\begin{document}

\title{Clustering-Based Weight Orthogonalization for Stabilizing Deep Reinforcement Learning
}


\author{
    Guoqing Ma $^{1,2,*}$, Yuhan Zhang $^{1,3,*}$, Yuming Dai $^{1,2}$, Guangfu Hao $^{1,3}$, Yang Chen $^{1,4}$, Shan Yu $^{1,2,4 }$ \dag
  \\
\thanks{* These authors contributed equally.}
\thanks{\dag Corresponding author: shan.yu@nlpr.ia.ac.cn} 
  $^{1}$ Institute of Automation, Chinese Academy of Sciences, Beijing, China\\
  $^{2}$ School of Future Technology, University of Chinese Academy of Sciences\\
  $^{3}$ School of Artificial Intelligence, University of Chinese Academy of Sciences, Beijing, China\\
  $^{4}$ Key Laboratory of Brain Cognition and Brain-inspired Intelligence Technology, Chinese Academy of Sciences\\
  \{maguoqing2022, zhangyuhan2022, daiyuming2025, haoguangfu2021, yang.chen\}@ia.ac.cn\\
}


\maketitle

\begin{abstract}
Reinforcement learning (RL) has made significant advancements, achieving superhuman performance in various tasks. However, RL agents often operate under the assumption of environmental stationarity, which poses a great challenge to learning efficiency since many environments are inherently non-stationary. This non-stationarity results in the requirement of millions of iterations, leading to low sample efficiency. To address this issue, we introduce the Clustering Orthogonal Weight Modified (COWM) layer, which can be integrated into the policy network of any RL algorithm and mitigate non-stationarity effectively. The COWM layer stabilizes the learning process by employing clustering techniques and a projection matrix. Our approach not only improves learning speed but also reduces gradient interference, thereby enhancing the overall learning efficiency. Empirically, the COWM outperforms state-of-the-art methods and achieves improvements of 9\% and 12.6\% in vision-based and state-based DMControl benchmark. It also shows robustness and generality across various algorithms and tasks.
\end{abstract}

\begin{IEEEkeywords}
Reinforcement learning, Non-Stationarity, Orthogonal weight modification
\end{IEEEkeywords}

\begin{figure*}[h]
  \centering
  \includegraphics[width =.95\linewidth]{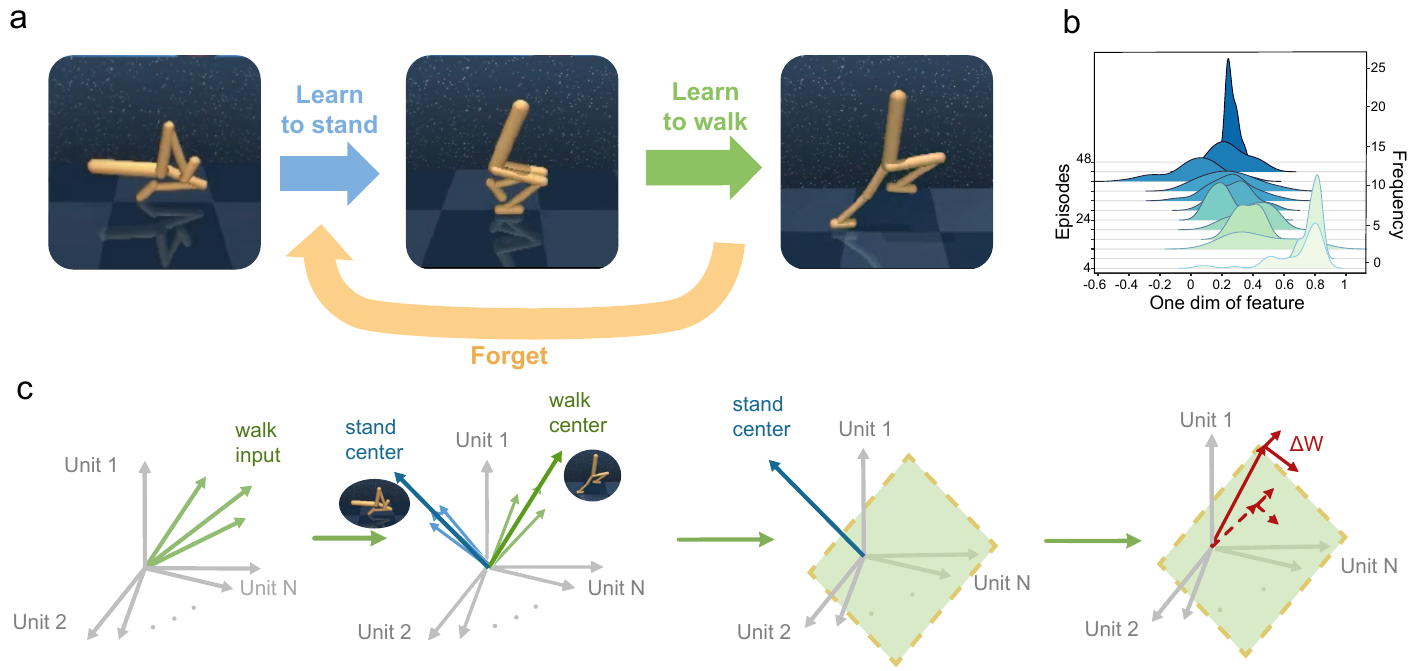}
  
  \caption{Non-stationarity in single-task reinforcement learning and its solutions. (a) Multiple sub-policies may exist within a single task. For example, in the walker walk task, the control of the robot can be divided into two sub-policies: learn to stand and learn to walk. (b) Distribution of state features over the course of training. (c) The COWM layer automatically identifies and protects the old policy. The gray arrows represent the input space for each layer of the neural network. The green and blue arrows indicate the input vectors during the learning process for the new and old policies, respectively. The green plane indicates the null space of the old policy. The red arrows show the gradient calculated by stochastic gradient descent for the new policy. The red dashed arrows represent the projection of the gradient into the null space.}
  \label{fig-1}
\end{figure*}

\section{Introduction}
In recent years, reinforcement learning (RL) has made significant progress across various domains, ranging from gaming to robotic control, often surpassing human performance \cite{silver2018general, mankowitz2023faster, fawzi2022discovering, kaufmann2023champion, ma2022improving, degrave2022magnetic}. Despite these advancements, a significant issue remains: the underlying assumption of a stationary environment \cite{puterman2014markov}. In numerous RL tasks, environments are inherently non-stationary \cite{xie2021deep,choi2001hidden}, with critical environmental components undergoing time-dependent changes. In extreme cases, the state transition function and reward function may both change over time \cite{khetarpal2022towards}. This non-stationarity poses a challenge for RL agents to adapt effectively to dynamic environments.

In a common RL scenario, multiple sequential tasks are encountered within a environment \cite{luketina2022meta,wilson2007multi}, leading to non-stationary state representations that pose challenges to the learning process. Additionally, even in environments with a single task, diverse sub-policies may need to be learned sequentially \cite{padakandla2020reinforcement}. This introduces non-stationarity in the state distribution \cite{khetarpal2022towards}. 
Such non-stationarity can lead to the problem of catastrophic forgetting in the field of continual learning \cite{kirkpatrick2017overcoming,lopez2017gradient}. Therefore, RL will also face catastrophic forgetting in general. However, it has not been thoroughly investigated in the field of RL. Catastrophic forgetting requires the network to continuously relearn the sub-policies. For instance, as shown in Figure \ref{fig-1}(a), in a single walk task, the agent should first learn how to stand and then learn how to walk. However, in the second phase, the agent forgets the standing skill, thereby necessitating a relearning process. This repetitive cycle leads to low learning efficiency.
However, traditional RL methods typically assume a fixed task distribution \cite{doshi2016hidden} and ignore the case of catastrophic forgetting, which is an inherent problem in neural networks. A promising path is to avoid interference between different sub-policies during the learning period. 

In this work, we introduce the COWM layer, designed to address the non-stationarity problem within a single task. Figure~\ref{fig-1}(b) illustrates the changes in state distribution during the agent's learning process, confirming the presence of state non-stationarity $\mathcal{S}(t)$. To address this, COWM layer employs clustering techniques and a projection matrix, enabling neural networks to learn new policies while minimizing interference with previously learned skills. As a result, the agent can potentially overcome non-stationarity, enhancing learning stability and speed. To maintain consistency, we refer to the RL algorithm that integrates the COWM layer as COWM.

The contributions of this paper can be summarized as follows:
\begin{itemize}

\item We demonstrate that non-stationarity in reinforcement learning occurs not only in multi-task scenarios but also in single-task ones. Specifically, we observe that in single-task settings, the state $\mathcal{S}(t)$ exhibits non-stationarity, which may result in reduced learning efficiency in the policy network. 

\item We propose the COWM layer which mitigates non-stationarity and enhances the stability of the policy network by constraining its gradients. This approach minimizes interference with previously learned skills when learning new policies, thereby improving sample efficiency and convergence speed.

\item The COWM layer exhibits high generalizability, making it suitable for all fully connected networks. It can be integrated into any RL algorithm that uses fully connected network as policy network.

\item Experiments show that our method outperforms state-of-the-art vision-based and state-based RL approaches, significantly improving performance across various classical control tasks.

\end{itemize}

\begin{figure}[ht]
  \centering
  \includegraphics[width =.9 \linewidth]{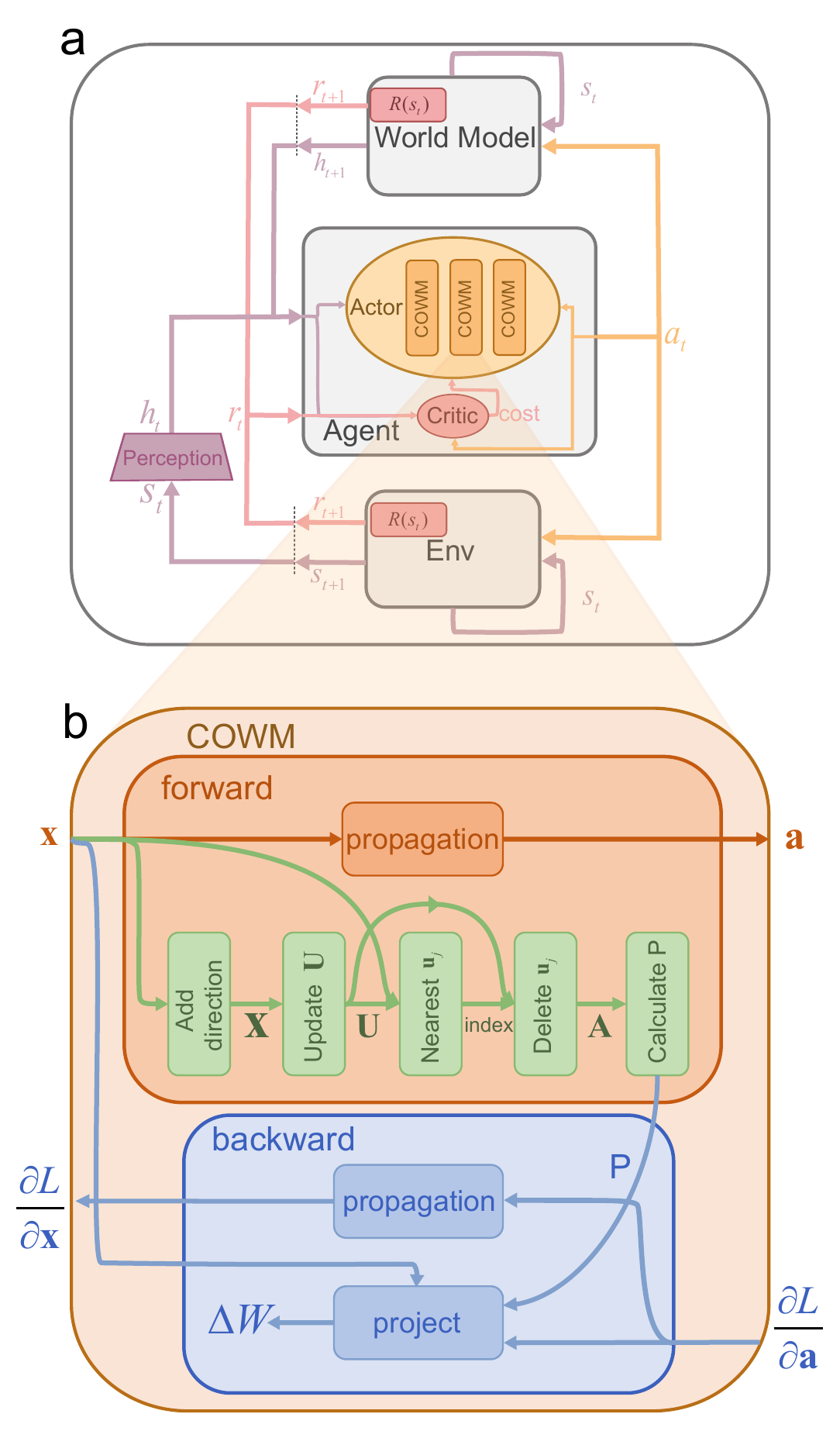}
  
  \caption{The COWM layer in the model-based reinforcement learning (MBRL) framework. (a) COWM replaces the linear layer within the actor, while other structures remain unchanged. (b) The internal structure and computational flow of the COWM layer. Computational flow is divided into two main processes: forward propagation (orange) and backward propagation (blue). The green section computes the projection matrix using historical input data. During backward propagation, the projection constrains the gradient using the projection matrix, resulting in the final weight updates.}%

  \label{fig-2}
\end{figure}

\section{Related Work}

\textbf{Non-stationarity in reinforcement learning.}
The fundamental components of the RL environment may exhibit time dependence \cite{khetarpal2022towards}. Considerable research efforts have been devoted to addressing the dynamics shift and learning generalizable policies for environments with changing dynamics.

One common practice is to train a context encoder \cite{lee2020context, chenlatent, zhou2018environment, wuimgfu, wu5033159generative} to associate the environment parameters. These context-based methods augment the state with an (latent) environment-related context \cite{lee2020context, yu2020learning, yupolicy, finn2017model}. A context is added to the policy input to help it adapt to a new environment \cite{chen2021offline}. They train context encoders to identify environment parameters. However, existing context-based methods need many interactions to recognize an environment context. 
To mitigate the problem of inefficient environment recognition, there are some attempts that leverage Importance Sampling (IS) \cite{eysenbachoff, liudara, niu2022trust}. Given the dynamics of the target environment, samples from the source environments are assigned with larger importance weights if they are more likely to happen in the target environment and vice versa. In sigle-task settings, IS is also used in PPO \cite{schulman2017proximal} to maintain stability. However, it is notoriously hard to balance the bias and variance when calculating the IS weights.


\textbf{Continual reinforcement learning.}
Our COWM layer avoids interfering with old policies when learning new ones, thereby mitigating the influence of representation non-stationarity. It aligns with several existing techniques in the field of Continual Reinforcement Learning (CRL). Below, we discuss related works that share similar objectives.

A primary approach to prevent catastrophic forgetting is to save an independent model for each task \cite{li2017learning, yoon2018lifelong, rosenfeld2018incremental, li2019learn, kanakis2020reparameterizing, abati2020conditional}. However, this approach is suboptimal due to the significant storage requirements and the need for task detection mechanisms. An alternative approach involves the use of regularization methods, such as employing Fisher information matrix to address catastrophic forgetting in RL \cite{kirkpatrick2017overcoming}. TRPO \cite{schulman2015trust} and PPO \cite{schulman2017proximal} also leverage the Fisher information matrix to mitigate instability in policy optimization. Furthermore, considering the constraints of MLPs, the use of a projection matrix can more effectively minimize learning overlap \cite{zeng2019continual}.
However, many continual learning methods require tasks to be learned sequentially and lack the ability to re-learn previous tasks \cite{ammar2014online, maurer2016benefit, d2020sharing, shi2021meta, yu2020gradient}. This limitation hinders their applicability in reinforcement learning scenarios where multiple sub-policies need to be learned concurrently. In contrast, our COWM layer supports repeated learning of old sub-policies while preserving the stability of each individual policy. 

Distillation and rehearsal-based methods are also commonly used in CRL. Knowledge distillation \cite{hinton2015distilling,rusu2015policy} involves using one neural network as a target for another to retain knowledge from past tasks and prevent forgetting. Similarly, experience replay \cite{lin1992self,pan2018organizing} reuses past experiences to maintain memory of old tasks during new tasks learning \cite{lillicrap2015continuous, barth2018distributed, atkinson2021pseudo}. However, compared to the COWM layer, these distillation and rehearsal-based approaches require explicit task sequence specification and require more parameters or storage. As a result, they are not suitable or efficient in single-task settings.


\section{COWM Layer in Policy network}

\subsection{Forward propagation and Projection matrix calculation}

We propose a novel COWM layer with a regularization method, which replaces the linear layer in the actor, as shown in Figure \ref{fig-2}(a). This approach preserves the original RL architecture without increasing the overall time complexity of the algorithm.

The core idea of our approach is to impose orthogonal constraints on the policy network's parameters $W$ during stochastic gradient descent optimization. These constraints enhance the stability of the policy network's learning process and improve data sample efficiency.

Specifically, for example, as shown in Figure \ref{fig-1}C, after learning how to stand, the agent begins to learn how to walk. In the actor network, each layer of the fully connected network receives a set of input data related to walking. The historical input data buffer contains input data related to standing. To prevent the new walking policy from interfering with the old standing policy, gradient updates must be projected onto the null space of the standing input data. A projection matrix is computed using the standing input data to achieve this. During actor training, gradients are projected using this matrix, resulting in a constrained gradient. It is then used to optimize the actor network, minimizing the impact on the standing policy.

In terms of the computational process, Figure \ref{fig-2}(b) illustrates the forward and backward propagation of the COWM layer. The forward propagation process remains unchanged (Eq. \ref{ec-2}).

Feed forward propagation: The forward propagation in the policy network is the same as in the classical fully connected linear network. The formula for the $l$-th layer is:

\begin{equation}\label{ec-2}
\begin{array}{l}
{{\bf{a}}_l} = W_l^T{{\bf{x}}_{l - 1}}\\
{{\bf{x}}_l} = f({{\bf{a}}_l}){\rm{ \ , \ }}l = 1,2,...,L,
\end{array} 
\end{equation}

\noindent Here, ${{\bf{x}}_{l - 1}} = \left[ {{x_1},{x_1},...,{x_b}} \right] \in {R^{d \times b}}$ represents the batch output data from layer $l-1$ and serves as the input data for layer $l$, where $b$ is batch size. ${W_l} \in {R^{d \times d}}$ is the weight matrix of the fully connected layer $l$. ${{\bf{a}}_l} \in {R^{d \times b}}$ is the result of linear transformation of input using the weight matrix. $f( \cdot )$ is activation function. The policy network consists of $L$ layers.

During each forward propagation process (Eq. \ref{ec-2}), a part of projection matrix $P_l$ is computed (Eq. \ref{ec-3}, \ref{ec-4}). $P_l$ is derived from historical input information, restricts the optimization space of the current weights (Eq. \ref{ec-5}).

Calculating the part of projection matrix $P_l$:

\begin{equation}\label{ec-3}
\begin{array}{l}{U_{l - 1}} = kmeans({X_{l - 1}},c)\\
{\bf{u}}_{l - 1}^j = nearest({\bf{x}}_{l - 1}^p,{U_{l - 1}})\\
{A_l} = \{ {U_{l - 1}}/{\bf{u}}_{l - 1}^j\}, \end{array}
\end{equation}

\begin{equation}\label{ec-4}
{P_l} = {A_l}{(A_l^T{A_l})^{ - 1}},
\end{equation}

\noindent where ${{\bf{X}}_{l - 1}} = \left[ {{\bf{\bar x}}_{l - 1}^1,{\bf{\bar x}}_{l - 1}^2,...,{\bf{\bar x}}_{l - 1}^F} \right] \in {R^{d \times F}}$ is a matrix composed of $F$ inputs. $F$ is the size of input direction buffer. ${\bf{\bar x}}_{l - 1}^f$ denotes the principal component of the $f$-th batch in layer $l$. Principal component ${{\bf{\bar x}}_{l - 1}^f}$ represents the normalized mean of the input of the $f$-th batch, defined as ${\bf{\bar x}}_{l - 1}^f = \frac{{\sum\limits_{j = 1}^b {{x_j}} }}{{\left| {\sum\limits_{j = 1}^b {{x_j}} } \right|}}$. The k-means clustering algorithm \cite{selim1984k} is employed to cluster $F$ directions into $c$ cluster centers, represented by ${U_{l - 1}} = [{\bf{u}}_{l - 1}^0,{\bf{u}}_{l - 1}^1,...,{\bf{u}}_{l - 1}^{\text{c}}]$. ${A_l} \in {{R}^{d \times s}}$ represents the preserved directions, which are the centers of all categories except the current one. 
$d$ denotes the number of neurons in the hidden layer. 
$s$ is the number of directions to be preserved, equivalent to the number of cluster centers minus one. The $i$-th column of ${A_l}$ represents the centers of the $i$-th category's input. However, this set of centers excludes the center of the current category $j$, denoted as $A_l^i = {\bf{u}}_{l - 1}^i,i \ne j$.
$P_l$ is part of the orthogonal projection matrix.

When the current input principal component is ${\bf{x}}_{l - 1}^f$, the nearest cluster center ${\bf{u}}_{l - 1}^j$ in ${U_{l - 1}}$ is found in terms of angular distance, and it is recorded as the $j$-th cluster center. By removing the $j$-th cluster center from all cluster centers, the final preserved directions ${A_l}$ are obtained. Through these operations, part of the orthogonal projection matrix $P_l$ is derived.

\subsection{Orthogonal weight modification and Backpropagation }

During backpropagation, the output layer of the neural network receives the gradient signal. On one hand, the gradient signal is propagated back to the input layer in the same manner as in standard BP algorithm. On the other hand, the input is projected using the projection matrix before calculating the weight updates.

Specifically, the weight updates are obtained using Equation (\ref{ec-5}), rather than the commonly used $\Delta W_l^{BP} =  - \eta \frac{{\partial L}}{{\partial {{\bf{a}}_l}}}{{\bf{x}}_{l - 1}}$ in stochastic gradient descent.

\begin{equation}\label{ec-5}
\Delta W_l^{COWM} =  - \eta \left( {\frac{{\partial L}}{{\partial {{\bf{a}}_l}}}{{\bf{x}}_{l - 1}} - \frac{{\partial L}}{{\partial {{\bf{a}}_l}}}{P_l}A_l^T{{\bf{x}}_{l - 1}}} \right),
\end{equation}

\begin{equation}\label{ec-52}
{W_l}(t + 1) = {W_l}(t) + \Delta W_l^{COWM},
\end{equation}

Here, $\Delta W_l^{COWM}$ represents the weight updates for the $l$-th layer obtained using COWM. $\eta $ is learning rate. $\frac{{\partial L}}{{\partial {{\bf{a}}_l}}}$ is the gradient of the loss with respect to the linear transformation result of layer $l$, obtained via backpropagation. ${{\bf{x}}_{l - 1}}$ is the input of layer $l$ and also the output from layer $l-1$. ${P_l}$ and ${A_l}$ are the parts of the projection matrix. It is important to note that ${P_l}$ is not a standard projection matrix in the traditional sense. It is just a part of the projection matrix. The Equation (\ref{ec-5}) is equivalent to the optimization using the standard projection matrix. However, compared to traditional projection methods, this approach significantly reduces computational complexity (see The time complexity impact of introducing COWM).

In the null space constrained by ${P_l}A_l^T$, the input-output mapping of old memories is maximally preserved $\pi :\left\{ {{S_{old}},{S_{new}}} \right\} \to \left\{ {{A_{old}},{A_{new}}} \right\}$. Without adding this constraint, when the network learns the current mapping $\pi :{S_{new}} \to {A_{new}}$, it cannot take into account other previously learned mappings, thereby interfering with the old mapping $\pi :{S_{old}}\bcancel{ \to }{A_{old}}$. Consequently, the agent easily forgets already learned policies, leading to the repetitive relearning of new and old policies, which in turn causes a decrease in learning efficiency and data utilization.

Finally, the COWM layer must also complete the backpropagation of gradients (Eq. \ref{ec-6}). The COWM layer transmits the gradient information from the output layer back to the input side in the same manner as a linear layer. This gradient is used to train the deep neural network layer by layer.

\begin{equation}\label{ec-6}
\frac{{\partial L}}{{\partial {{\mathbf{x}}_{l - 1}}}} = {W_l}\frac{{\partial L}}{{\partial {{\mathbf{a}}_l}}},
\end{equation}

We adopt the actor-critic learning settings with a model-based framework. 


\subsection{Proof of memory protection for COWM}

COWM retains the memory of old policies while learning new ones. It satisfies $\pi :\left\{ {{\mathcal{S}_{old}},{\mathcal{S}_{new}}} \right\} \to \left\{ {{\mathcal{A}_{old}},{\mathcal{A}_{new}}} \right\}$ simultaneously. In this section, we will explain the mathematical principle behind why a neural network layer tends to forget old memories and why COWM preserves these old memories.

Considering a simple case. The algorithm has to complete two policies $\mathop {{\pi}}\limits_{s \sim {\mathcal{S}_{1}}} \left( {{W^*}s} \right) \sim {\mathcal{A}_{1}}$, $\mathop {{\pi }}\limits_{s \sim {\mathcal{S}_{2}}} \left( {{W^*}s} \right) \sim {\mathcal{A}_{2}}$


\begin{equation}
\begin{gathered}
  {W^*} = \arg \mathop {\max }\limits_W E\left[ {\sum\limits_{i = 0}^T {{\gamma ^{t + i}}\mathop R\limits_{s \sim {S_{1}}} \left( {{s_{t + i}},\pi \left( {W{s_{t + i}}} \right)} \right)} } \right. \hfill \\
  \left. { + \sum\limits_{i = 0}^T {{\gamma ^{t + i}}\mathop R\limits_{s \sim {S_{2}}} \left( {{s_{t + i}},\pi \left( {W{s_{t + i}}} \right)} \right)} } \right] \hfill ,\\ 
\end{gathered} 
\end{equation}

In reinforcement learning, the non-stationarity of states often makes the learning process like a sequence learning task (Figure \ref{fig-1}a, b). To simplify the problem, we focus on learning two policies sequentially: first learning policy 1, followed by learning policy 2.

\begin{figure*}[h]
  \centering
  \includegraphics[width =.9 \linewidth]{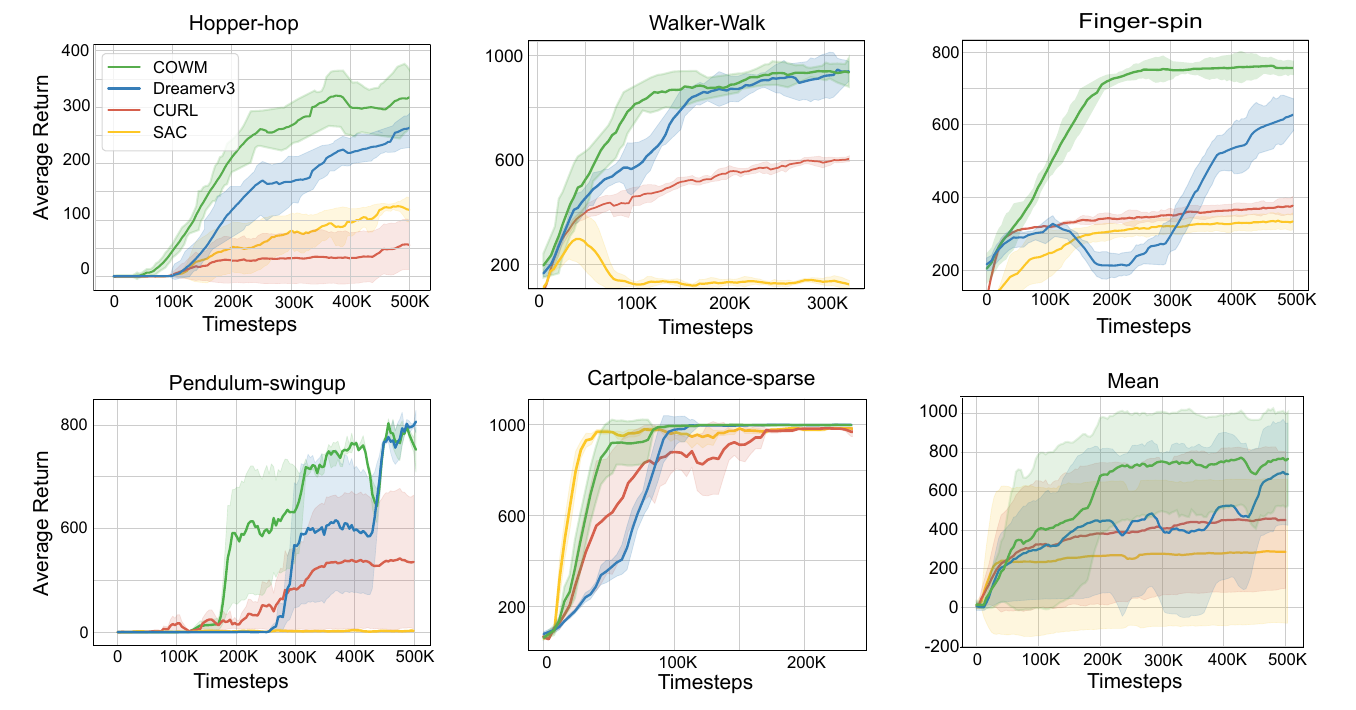}
  \caption{Training curves for 5 tasks in vision-based DMControl. SAC, CURL and DreamerV3 are compared. We replicated the results of SAC, CURL, and DreamerV3 in the same environments on NVIDIA A40 GPU. Our method is represented in green, while DreamerV3 is in blue, CURL is in red and SAC is in yellow. Mean and 95\% CIs over 3 seeds.}
  \label{fig-3}
\end{figure*}

Policy 1: 
\begin{equation}
{W^{*1}} = \arg \mathop {\max }\limits_W E\left[ {\sum\limits_{i = 0}^T {{\gamma ^{t + i}}\mathop R\limits_{s \sim {S_{1}}} \left( {{s_{t + i}},\pi \left( {W{s_{t + i}}} \right)} \right)} } \right],
\end{equation}

\begin{equation}
\mathop \pi \limits_{s \sim {S_1}} \left( {{W^{*1}}s} \right) \sim {\mathcal{A}_{1}},
\end{equation}

In a conventional neural network layer, it tends to forget old memory when learning policy 2.

Policy 2: 
\begin{equation}
{W^{*2}} = \arg \mathop {\max }\limits_W E\left[ {\sum\limits_{i = 0}^T {{\gamma ^{t + i}}\mathop R\limits_{s \sim {S_{2}}} \left( {{s_{t + i}},\pi \left( {W{s_{t + i}}} \right)} \right)} } \right],
\end{equation}

\begin{equation}
W(t + 1)\mathop  = \limits_{s \sim {S_2}} W(t) - \alpha \frac{{\partial E}}{{\partial W}} = W(t) - \alpha \frac{{\partial L}}{{\partial \left( {Ws} \right)}}{s^T},
\end{equation}

let ${g_2}\mathop  = \limits_{s \sim {S_2}} \sum\limits_{t = 1}^T {\frac{{\partial L}}{{\partial \left( {Ws} \right)}}}  $, this gives
\begin{equation}
W_{BP}^{*2}\mathop  = \limits_{s \sim {S_2}} {W^{*1}} - \alpha {g_2}{s^T},
\end{equation}

as a result,

\begin{equation}
\begin{gathered}
  \mathop \pi \limits_{{s_2} \sim {S_2}} \left( {W_{BP}^{*2}{s_2}} \right) \sim {\mathcal{A}_2} \hfill \\
  \mathop \pi \limits_{{s_1} \sim {S_1}} \left( {W_{BP}^{*2}{s_1}} \right) = \mathop \pi \limits_{{s_1},{s_2} \sim {S_1},{S_2}} \left( {\left( {{W^{*1}} - \alpha {g_2}s_2^T} \right){s_1}} \right) \hfill \\
   = \mathop \pi \limits_{{s_1},{s_2} \sim {S_1},{S_2}} \left( {{W^{*1}}{s_1} - \alpha {g_2}s_2^T{s_1}} \right){\cancel{ \sim }}{\mathcal{A}_1} \hfill ,\\ 
\end{gathered} 
\end{equation}

In general, ${{s_1}}$ and ${{s_2}}$ are not orthogonal, and ${g_2} \ne 0$, so $\alpha {g_2}s_2^T{s_1} \ne 0$.
$\mathop \pi \limits_{{s_1} \sim {S_1}} \left( {W_{BP}^{*2}{s_1}} \right)\cancel{ \sim }{\mathcal{A}_1}$ means the linear layer with BP forgets old memory.
\begin{equation}
{\pi _{W_{BP}^{*2}}}:\left\{ {\begin{array}{*{20}{c}}
  {{S_1}\cancel{ \to }{\mathcal{A}_1}} \\ 
  {{S_2} \to {\mathcal{A}_2}} 
\end{array}} \right\},
\end{equation}

In the COWM layer, a projection matrix is calculated and constraints the gradient when learning policy 2 while protecting old memory. A standard projection matrix is calculated as,
\begin{equation}
P = I - A{\left( {{A^T}A} \right)^{ - 1}}{A^T},
\end{equation}

\begin{figure*}[h]
  \centering
  \includegraphics[width =1. \linewidth]{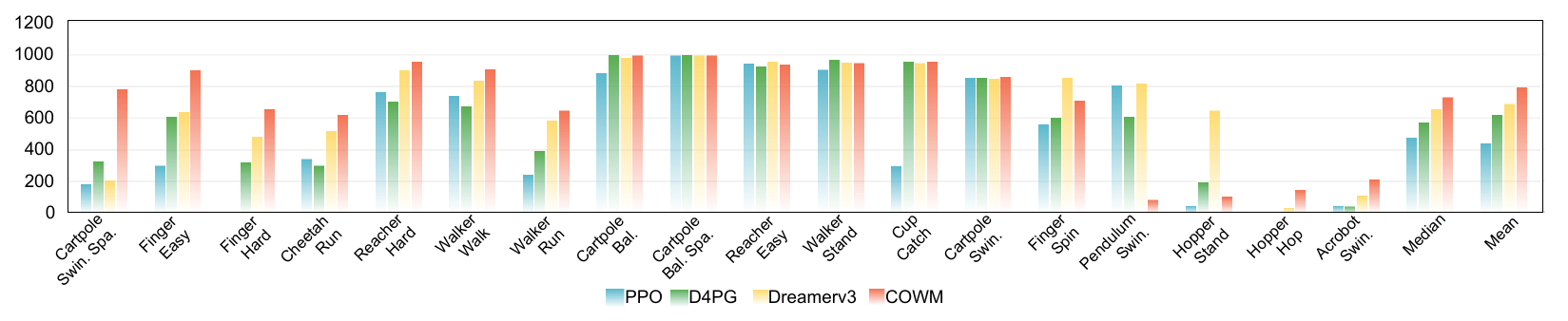}
  \caption{Performance comparison to the existing methods in state-based DMControl. It contains rewards for 18 tasks under 250K interactions with the environment. For COWM, COWM layers are implemented without any hyperparameter fine-tuning compared with vision-based COWM showing its generalization ability. PPO is regularization based method and D4PG is rehearsal based method to mitigate non-stationarity of DRL.} 
  \label{fig-7}
\end{figure*}

In the main text, we select two cluster centers and remove one of them, leaving only a single cluster center. This remaining cluster center can be considered as the mean of $S_1$. That is $A = {{\bar s}_1}$. For OWM when learning policy 2, the update formula is,
\begin{equation}
W(t + 1)\mathop  = \limits_{{s_2} \sim {S_2}} W(t) - \alpha \frac{{\partial L}}{{\partial W}}P = W(t) - \alpha \frac{{\partial L}}{{\partial \left( {W{s_2}} \right)}}s_2^TP,
\end{equation}

 which is,

 \begin{equation}
W_{COWM}^{*2}\mathop  = \limits_{{s_2} \sim {S_2}} {W^{*1}} - \alpha {g_2}s_2^TP,
\end{equation}

as a result,

\begin{equation}
\begin{gathered}
  \mathop \pi \limits_{{s_2} \sim {S_2}} \left( {W_{COWM}^{*2}{s_2}} \right) \sim {\mathcal{A}_2} \hfill \\
  \mathop \pi \limits_{{s_1} \sim {S_1}} \left( {W_{COWM}^{*2}{s_1}} \right) = \mathop \pi \limits_{{s_1},{s_2} \sim {S_1},{S_2}} \left( {\left( {{W^{*1}} - \alpha {g_2}s_2^TP} \right){s_1}} \right) \hfill \\
   = \mathop \pi \limits_{{s_1},{s_2} \sim {S_1},{S_2}} \left( {{W^{*1}}{s_1} - \alpha {g_2}s_2^TP{s_1}} \right) \sim {\mathcal{A}_1} \hfill ,\\ 
\end{gathered} 
\end{equation}

\noindent because of $E\left[ {\alpha {g_2}s_2^TP{s_1}} \right] = \alpha {g_2}\bar s_2^TP{{\bar s}_1} = 0$. The results means that COWM layer protects old memory while learning new policy.

\begin{equation}
{\pi _{W_{COWM}^{*2}}}:\left\{ {\begin{array}{*{20}{c}}
  {{S_1} \to {\mathcal{A}_1}} \\ 
  {{S_2} \to {\mathcal{A}_2}} 
\end{array}} \right\},
\end{equation}


\begin{table}[h]
  \caption{Performance of COWM}
   Comparison in vision-based DMControl tasks under 1M environment steps across 3 random seeds. The bold values are the highest among each row.
  \centering
  \setlength{\tabcolsep}{3pt} 
  \begin{tabular}{c|cccc|c}
    \toprule
    \textbf{Task}     & \textbf{SAC}     & \textbf{CURL} & \textbf{DrQ-v2} & \small{\textbf{DreamerV3}} & \textbf{COWM} \\
    \midrule
    \midrule
    Acrobot Swingup & 5.1  & 5.1 & 128.4 & 210  & \textbf{322} \\
    Cartpole Balance  & \textbf{963.1} & \textbf{979} & \textbf{991.5} & \textbf{996.4} & \textbf{999.7}    \\
    Cartpole Sparse     & \textbf{950.8} & \textbf{981} & \textbf{996.2} & \textbf{1000} & \textbf{1000} \\
    Cartpole Swingup & 692.1 & 762.7 & \textbf{858.9} & 819.1 & 831.4 \\
    Cartpole Sparse & 154.6 & 236.2 & 706.9 & \textbf{792.9} & 770.1 \\
    Cheetah Run & 27.2 & 474.3 & 691 & 728.7 & \textbf{866.1} \\
    Cup Catch & 163.9 & \textbf{965.5} & 931.8 & 957.1 & \textbf{983.4} \\
    Finger Spin & 312.2 & \textbf{877.1} & \textbf{846.7} & 818.5 & 829.3 \\
    Finger Turn Easy & 176.7 & 338 & 448.4 & 787.7 & \textbf{969.7} \\
    Finger Turn Hard & 70.5 & 215.6 & 220 & 810.8 & \textbf{942.2} \\
    Hopper Hop & 3.1 & 152.5 & 189.9 & 369.6 & \textbf{474.5} \\
    Hopper Stand & 5.2 & 786.8 & 893 & 900.6 & \textbf{956.8} \\
    Pendu. Swingup & 560.1 & 376.4 & 839.7 & 806.3 & \textbf{910.1} \\
    Quadruped Run & 50.5 & 141.5 & 407 & 352.3 & \textbf{426.2} \\
    Quadruped Walk & 49.7 & 123.7 & \textbf{660.3} & 352.6 & 415.6 \\
    Reacher Easy & 86.5 & 609.3 & 910.2 & 898.9 & \textbf{985.1} \\
    Walker Run & 26.9 & 376.2 & 517.1 & 757.8 & \textbf{764.4} \\
    Walker Stand & 159.3 & 463.5 & 974.1 & 976.7 &\textbf{ 996.9} \\
    Walker Walk & 38.9 & 828.8 & 762.9 & 955.8 & \textbf{982.8} \\
    \midrule
    \rowcolor{gray!20}
    Median & 86.5 & 463.5 & 762.9 & 810.8 &  \textbf{942.2} \\
    \rowcolor{gray!20}
    Mean & 236.7 & 510.2 & 682.8 & 752.2 &  \textbf{822.9} \\
    \bottomrule
    
  \end{tabular} 
   \begin{minipage}{\linewidth}
  \raggedright
  \end{minipage}
  \label{table3}
\end{table}

\section{Experiments}
We evaluate the performance of COWM on the widely-used DMControl benchmark \cite{tassa2018deepmind}, measuring both data-efficiency and overall performance of our method and baselines at various environment steps. 

\textbf{RL baselines:} We compared several baselines for continuous control: 1) For vision-based RL, the comparison baselines include the model-based method DreamerV3, and data-augmentation-based method CURL \cite{laskin2020curl} and well-used baseline SAC \cite{haarnoja2018soft}.  2) For state-based RL, the comparison baselines include the regularization-based method PPO, the rehearsal-based D4PG \cite{barth2018distributed} and the model-based method DreamerV3. Specifically, PPO uses KL divergence as a constraint, where its second derivative forms the Fisher information matrix. This approach aligns with EWC \cite{kirkpatrick2017overcoming}, which also uses the Fisher information matrix to mitigate non-stationarity in RL. D4PG, on the other hand, employs replay buffer to mitigate the catastrophic forgetting in neural networks.

\textbf{DMControl suite:} We benchmark the performance of COWM in continuous action control environments \cite{escontrela2024video, grossman2024differentially, metcalf2024can}, which are typically more challenging than discrete action environments \cite{li2023efficient, aleksandrowicz2023metrics}. Specifically, we focus on the DMControl suite, which consists of two categories: vision-based and state-based. 

\subsection{Results on DMControl}

We evaluate the methods on 19 vision-based DMControl tasks. All experimental results and part of the training curves are presented in Table~\ref{table3} and Figure \ref{fig-3}, respectively. The results show that COWM outperforms previous SOTA methods under 1M environment interactions, including DreamerV3 on 15 tasks.
On average, COWM achieves approximately 9\% improvement over DreamerV3. Furthermore, as shown in Figure \ref{fig-3}, COWM demonstrates improvement in both training stability and sample efficiency, yielding better performance when compared to other algorithms. The reward of COWM starts increasing from the very beginning in each environment. 
COWM's reward increases at a faster rate than other algorithms, requiring fewer interactions to reach optimal performance. Additionally, in environments such as Hopper-hop, Finger-spin, and Walker-walk, COWM achieves higher maximum rewards compared to other methods. This suggests  that by enhancing the stability of policy learning, COWM not only improves sample efficiency but also helps the agent learn more optimal policies.

\begin{figure*}[h]
  \centering
  \includegraphics[width =1.\linewidth]{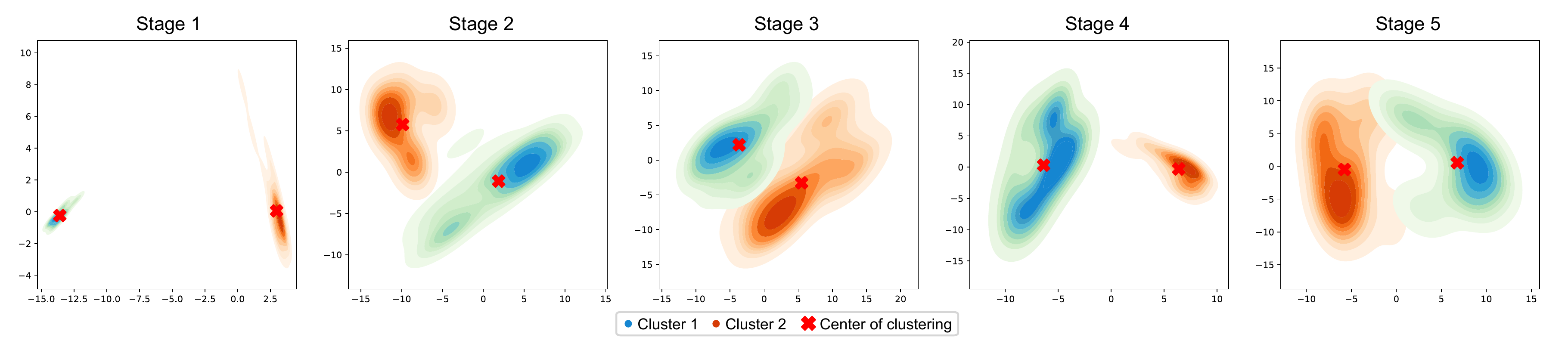}
  \caption{During training with the COWM layer, the representations in the actor network are clustered. There are two cluster centers, meaning the representations are divided into two categories, represented by blue and orange areas. The cluster centers are marked by red crosses.}
  \label{fig-cluster}
\end{figure*}

We also evaluate the methods on 18 state-based DMControl tasks to verify its generalization ability. We use the same hyperparameters of COWM layer for vision-based DMControl tasks and implement it into state-based DMControl tasks directly without any fine-tuning. All experimental results are shown in Figure \ref{fig-7}. 
The experimental results show that COWM outperforms previous SOTA methods under 250K interactions with the environment on 11 tasks, especially on "Cartpole SwingupSparse" and "Finger TurnEasy" tasks. On average, there is an improvement of approximately 12.6\% as compared to DreamerV3. 
State-based DMControl experiment indicates that COWM layer not only improves sample efficiency but also has generalization ability.

\subsection{Analysis}

We extract the representation data from the hidden layers of the actor network and visualize the clustering situation of the COWM layer. As shown in Figure \ref{fig-cluster}, the representations were divided into five stages, arranged sequentially from early to final stages. We set the number of cluster centers $c=2$ to cluster the representations into two categories. The results show that the hidden layer representations are asymmetric and uneven in space at early stages, especially in stage 2 and stage 3. It is not favorable for the stable learning of the two sub-policies. In the later stages of training, the hidden layer representations exhibit a trend of differentiation into two categories. The distribution area in each category becomes more balanced. The cluster centers also evenly divide the state space. This indicates that, on one hand, the COWM layer causes the actor network to have a clustering trend in representations, which means different policies are located in different representation spaces. On the other hand, the cluster centers accurately identify the main directions of different policies, effectively protecting old policies with the help of the orthogonal weight modification strategy.

\begin{figure}[h]
  \centering
  \includegraphics[width =1. \linewidth]{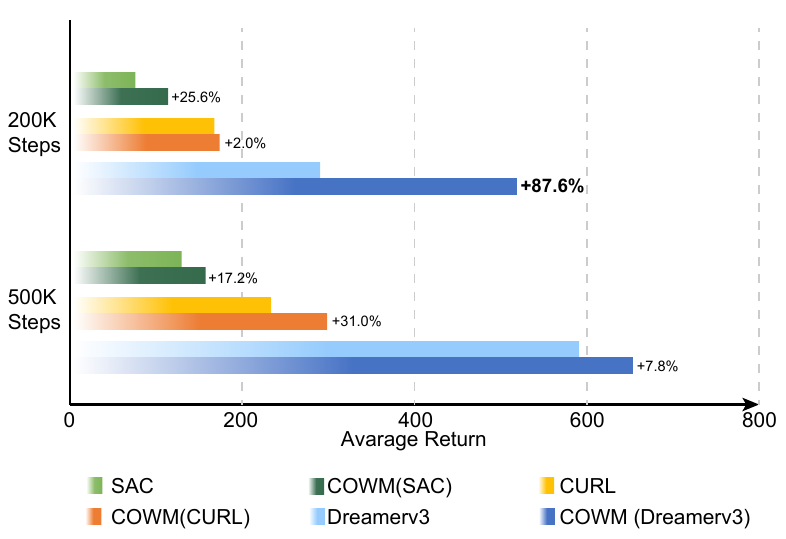}
  
  \caption{Generalizability of COWM across various RL algorithms. Comparison of DreamerV3-based, CURL-based and SAC-based COWM against their base algorithms (DreamerV3 CURL and SAC). Bars of the same color family, i.e., green, orange and blue, represent a group of algorithms with the same RL framework.}
  \label{fig-6}
\end{figure}

\section{Ablation studies}

In this section, we conduct various ablation studies on continuous control tasks to determine the generalizability and effectiveness of COWM across various RL algorithms and the factors affecting the performance, including different numbers of cluster centers and cluster iterations.

\subsection{Generalizability of COWM across various RL algorithms}

We integrate the COWM layer into various RL algorithms, including SAC, CURL, and DreamerV3, to evaluate its generalizability by comparing the changes in rewards before and after introducing the COWM layer. No algorithm-specific hyperparameter tuning is conducted, all hyperparameters for COWM remain consistent across different tasks. The experiments are conducted in three vision-based DMControl suite tasks \cite{wu2024image}. The average rewards across these tasks are recorded for analysis. 
Figure~\ref{fig-6} shows that after 200K steps of iteration, SAC and DreamerV3 have significant improvements with the COWM layer, indicating that COWM enhances both the learning speed and sample efficiency of these algorithms. After 500K steps of iteration, the results demonstrate that incorporating COWM improves final performance across all three algorithms, suggesting that COWM possesses cross-algorithm generalizability.

\begin{figure}[h]
  \centering
  \includegraphics[width =1.\linewidth]{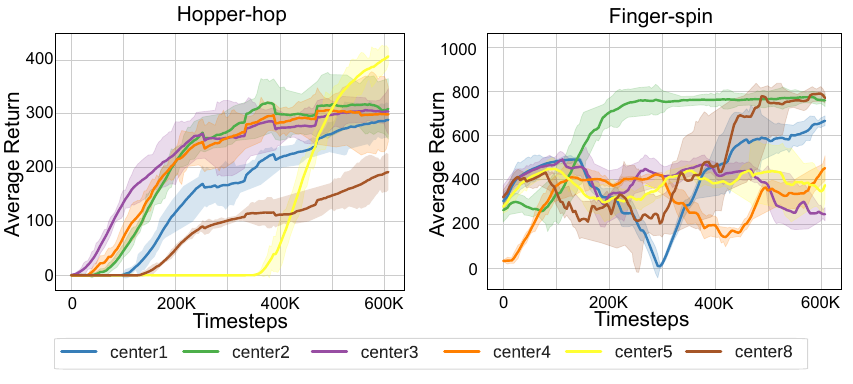}
  
  \caption{The effect of the different hyperparameters $c$ with $k=10$ in the COWM policy learning. Mean and 95\% CIs over 3 seeds.}
  \label{fig-4}
\end{figure}

\subsection{The effect of the number of cluster centers and cluster iterations}

The hyperparameter $c$ represents the number of cluster centers, and $k$ represents the number of cluster iterations. Together, they control the impact of the COWM layer on policy learning. In the first experiment, we set $k = 10$ and investigate the impact of different $c$ on policy performance. $c$ determines the number of policies to protect. The experimental results are shown in Figure~\ref{fig-4}. According to the results of two different tasks, an appropriate $c$ is crucial for obtaining better performance. If $c$ is too small or too large, it will cause performance degradation. 
Theoretically, different tasks have different numbers of relatively independent policies. However, based on our experiments on various tasks in the DMControl suite, setting $c=2$ generally achieves better performance, balancing efficiency and stability. This suggests that in relatively simple tasks, dividing the policy into two parts helps improve the efficiency of policy learning.

\begin{figure}[h]
  \centering
  \includegraphics[width =1.\linewidth]{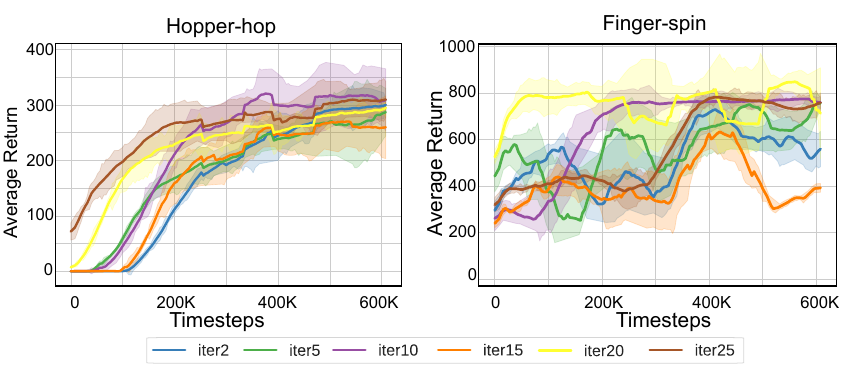}
  
  \caption{The effect of the different hyperparameters $k$ with $c=2$ in the COWM policy learning.Mean and 95\% CIs over 3 seeds.}
  \label{fig-5}
\end{figure}

In the second experiment, we set $c$ = 2 and investigate the impact of different $k$ on policy performance. $k$ is the number of iterations for calculating clustering centers using k-means. The results are shown in Figure~\ref{fig-5}. The results of two different tasks indicate that when the $k$ is too small, the agent performs at a relatively low level. This indicates that too few iterations cannot ensure that the cluster centers correctly fall into the dense regions of the representations, thereby affecting the COWM layer's ability to correctly protect old policies. When $k$ is relatively large, apart from some unstable experimental results, the convergence speed is faster than those with $k=2$. This suggests that more clustering iterations help the COWM layer find the optimal direction of policy representation. However, too many iterations are unnecessary. When the number of iterations exceeds 10, more iterations does not significantly improve the performance.

\section{Conclusion}

In conclusion, our proposed COWM layer effectively addresses the non-stationarity challenge in reinforcement learning by introducing clustering orthogonal constraints into the policy network. This approach enhances the stability of the policy network and improves both sample efficiency and convergence speed. COWM's broad applicability to fully connected networks makes it a general solution for RL algorithms. Our experimental results demonstrate that COWM improves performance by 87.6\% over DreamerV3 after 200k interactions, showcasing its high data efficiency. It also outperforms state-of-the-art RL methods, achieving improvements of 9\% and 12.6\% in vision-based and state-based DMControl tasks, demonstrating significant gains in sample efficiency across multiple benchmarks. This work demonstrates the importance of mitigating state non-stationarity. We hope COWM can be applied to a wider range of non-stationary data scenarios.


\section*{Acknowledgment}
This work was supported in part by the Strategic Priority Research Program of the Chinese Academy of Sciences (CAS)(XDB1010302), CAS Project for Young Scientists in Basic Research, Grant No. YSBR-041 and the International Partnership Program of the Chinese Academy of Sciences (CAS) (173211KYSB20200021).

\bibliographystyle{IEEEtran} 
\bibliography{IEEEexample}

\end{document}